\pdfoutput=1
\documentclass[journal]{IEEEtran}
\usepackage{amsmath,amsfonts}
\usepackage{algorithmic}
\usepackage{algorithm}
\usepackage{array}
\usepackage[caption=false]{subfig}
\usepackage{textcomp}
\usepackage{stfloats}
\usepackage{url}
\usepackage{verbatim}
\usepackage{graphicx}
\usepackage{booktabs,tabularx}
\usepackage{cite}
\usepackage{stackengine,xcolor,enumitem}
\makeatletter
\newcommand\HUGE{\@setfontsize\Huge{50}{60}}
\makeatother    
\DeclareRobustCommand\circnum[1]{%
  \stackinset{c}{}{c}{.1ex}{\small\textcolor{white}{#1}}%
  {\abovebaseline[-.7ex]{\HUGE{$\bullet$}}}%
}

\definecolor{listinggray}{gray}{0.95}
\usepackage{listings}

\usepackage{tikz}

\newcommand\submittedtext{%
  \footnotesize This work has been submitted to the IEEE for possible publication. Copyright may be transferred without notice, after which this version may no longer be accessible.}

\newcommand\submittednotice{%
\begin{tikzpicture}[remember picture,overlay]
\node[anchor=south,yshift=10pt] at (current page.south) {\fbox{\parbox{\dimexpr0.65\textwidth-\fboxsep-\fboxrule\relax}{\submittedtext}}};
\end{tikzpicture}%
}
\begin{document}

\title{An Agentic AI Framework for Training General Practitioner Student Skills}
\author{Victor De Marez, Jens Van Nooten, Luna De Bruyne, and Walter Daelemans
\thanks{Manuscript received December 20, 2025. This research received funding from the Flemish Government under the ``Onderzoeksprogramma Artificiële Intelligentie (AI) Vlaanderen'' programme.
\textit{(Corresponding author: Victor De Marez.)}}%
\thanks{Victor De Marez, Jens Van Nooten, Luna De Bruyne and Walter Daelemans are with the Center for Computational Linguistics, Psycholinguistics and Sociolinguistics (CLiPS), University of Antwerp, Antwerp, Belgium (e-mail: firstname.lastname@uantwerpen.be).}
\thanks{This article has supplementary downloadable material available at https://doi.org/...}}

\maketitle
\submittednotice

\begin{abstract}
Advancements in large language models offer strong potential for enhancing virtual simulated patients (VSPs) in medical education by providing scalable alternatives to resource-intensive traditional methods. However, current VSPs often struggle with medical accuracy, consistent roleplaying, scenario generation for VSP use, and educationally structured feedback. We introduce an agentic framework for training general practitioner student skills that unifies (i) configurable, evidence-based vignette generation, (ii) controlled persona-driven patient dialogue with optional retrieval grounding, and (iii) standards-based assessment and feedback for both communication and clinical reasoning. We instantiate the framework in an interactive spoken consultation setting and evaluate it with medical students ($\mathbf{N{=}14}$). Participants reported realistic and vignette-faithful dialogue, appropriate difficulty calibration, a stable personality signal, and highly useful example-rich feedback, alongside excellent overall usability. These results support agentic separation of scenario control, interaction control, and standards-based assessment as a practical pattern for building dependable and pedagogically valuable VSP training tools.
\end{abstract}

\section{Introduction}
\noindent\IEEEPARstart{I}{n} medical education worldwide, simulated patients (SP), which are trained actors who portray patients with predefined symptoms and behaviors \cite{CHURCHOUSE2012e363}, are used to teach essential skills such as history taking and communication skills, and explaining a diagnosis. They are also a traditional part of the \textit{Objective structured clinical exams} (OSCE) assessment of students, in which they are meant to focus on a specific skill, so that SPs are used to systematically measure clinical and communication skills in a standardized way \cite{Cleland01012009}. 

\begin{figure}
    \centering
    \includegraphics[width=0.8\linewidth]{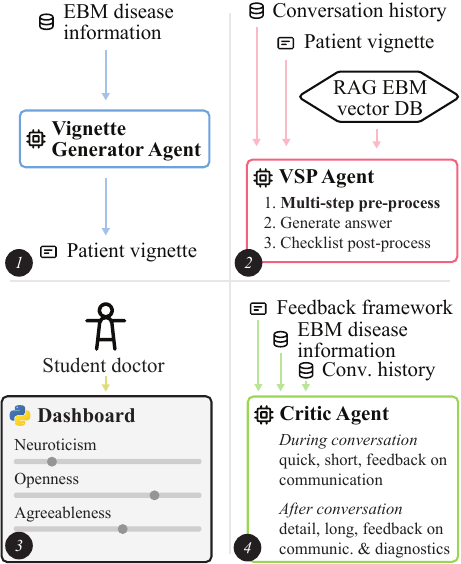}
    \caption{Core schema of the four main contributions of our framework: clinical vignette generation, a three-step VSP generation method, personality customization, and automated feedback generation}.
    \label{fig:maincontributions}
\end{figure}

However, training and hiring these qualified SPs is a procedure that requires substantial investments in resources and time \cite{rideout2001transforming}. Furthermore, despite rigorous training efforts, perfect replicability of the scenario is impossible due to inherent human variance and errors. Additionally, the educational setting with an SP can be distracting and stressful due to the presence of the tutor and other students \cite{rideout2001transforming}.

Virtual Simulated Patients (VSPs) are computer simulations of real-life patients programmed with clinical vignettes (clinical scenarios that include patient information) that allow a learner to obtain a medical history, make a diagnosis, and prescribe a treatment plan \cite{cook2009virtual}. Initially, VSPs were costly, and limited in realism, natural language capabilities, effectiveness and applicability. However, advances in artificial intelligence have accelerated development of VSPs \cite{hamilton2024evolution, cook2025creating}, thereby offering effective solutions to the aforementioned problems of SPs. Despite their initial limitations, VSPs offer multiple advantages over SPs. For instance, virtual patients can be used by unlimited learners at virtually no incremental cost, therefore being more cost-friendly and less resource-intensive \cite{hayes2009using}. This scalability allows for interaction beyond stressful educational settings, for instance from home. Moreover, VSPs can be configured to follow a predefined vignette consistently and to incorporate a larger number of case details than is typically feasible for human actors.

Early VSPs were mainly comprised of rule-based backends, as in \cite{Maicher2017ConversationalVSP}. More recently, large language models (LLMs) such as GPT-4o \cite{hurst2024gpt} have largely replaced rule-based systems due to their ability to provide human-like responses. Due to their long context lengths and contextual understanding, LLMs are shown to be promising for creating realistic and real person-based agents with various personalities for simulating personas. For this reason, they are already used in various educational fields \cite{wang2025evaluating, Wang2025BeyondPF, chu2025llm}. Nonetheless, LLMs come with several challenges for educational and agentic contexts, complicating their integration into VSPs. For example, a hallucinating LLM can give plausible but inaccurate information, whether it is incorrect information relative to a VSP's scenario or inaccurate feedback in a critic agent \cite{chu2025llm}. If these hallucinations result in medical incorrectness, despite the vast medical knowledge in these LLMs, agents become unreliable \cite{perets2025cupcase}. Additionally, bias in large language models can lead to symptoms or conditions being linked to certain stereotypes more often, making students unprepared for real-world use cases \cite{bakkum2024using}.

The intersection of VSPs and LLMs leads to challenges of their own kind. Without thorough prompting and control, a roleplaying LLM tends to lose its verbal communication style beyond 4--6 turns, shifting back to its initial agreeable \textit{helpful assistant} tone \cite{bodonhelyi2025beyond}. Additional difficulty comes from the side of the end users. For instance, students tend to dislike a VSP more when it is unrealistic, limited in natural responses, repetitive, or when the task is too difficult \cite{kelly2022scoping}. The latter can be the case if VSP objectives are not aligned with educational goals in medical training, which is often the case in current VSPs due to a lack of educational frameworks \cite{Bowers_Graydon_Ryan_Lau_Tomlin_2024,meynhardt2025advanced}. Finally, the clinical vignettes in VSPs are often real cases or developed by experts, while many different vignettes are needed to prevent fraud, for example when used as an assessment tool during student exams \cite{Bowers_Graydon_Ryan_Lau_Tomlin_2024, bakkum2024using}.

These challenges motivate a VSP design that prioritizes (i) \textbf{scenario diversity}, (ii) \textbf{scenario fidelity} (patient facts according to or consistent with the vignette), (iii) \textbf{medical grounding} when clinical knowledge is required, (iv) \textbf{stable persona expression} across turns, and (v) \textbf{standards-based assessment} that produces actionable feedback aligned with educational rubrics rather than generic chatbot advice.
 
To address these challenges, we present an AI framework for training general practitioner student skills. Our main contributions, as visualized in Fig. \ref{fig:maincontributions}, are the following:
\begin{enumerate}[label=\protect\circnum{\theenumi}]
\item \textbf{Scenario control:} generation of consistent, medically grounded VSP scenarios using evidence-based medicine (EBM) and LLMs, enabling configurable training cases at scale;
\item \textbf{Behavior reliability:} a response control scheme that improves \textit{scenario fidelity} and reduces hallucinations by separating (a) structured reasoning over the utterance type, (b) optional retrieval augmentation when needed, and (c) a final constraint/cleanup step prior to delivery;
\item \textbf{Persona variation:} customization of VSP interaction style via Big Five traits (openness, conscientiousness, extraversion, agreeableness, neuroticism), operationalized through prompt-based control;
\item \textbf{Standards-based assessment:} automated feedback on student communication based on the Master Interview Rating Scale \cite{stillman1976use} (25 criteria) and on diagnostic performance by comparing student actions against evidence-based guidelines.
\end{enumerate}

The remainder of the paper is structured as follows. Section \ref{sec:relatedwork} reviews recent related work. In Section \ref{sec:sysarch}, the framework is described with its three agent roles: (i) a scenario generator, (ii) a conversational agent, and (iii) a standards-based critic. Finally, in Section \ref{sec:evaluation}, we evaluate the framework in an interactive spoken consultation setting.

\section{Related Work}\label{sec:relatedwork}
\subsection{LLMs for VSPs}
\noindent The use of Virtual Simulated Patients (VSPs) in medical education predates recent advances in large language models (LLMs), with reviews highlighting their potential but also limitations in realism, curricular integration, and particularly feedback generation \cite{kelly2022scoping, lee_effective_2020, bowers_artificial_2024}. Early AI-driven VSPs relied on rule-based systems or simple NLP techniques, often struggling with conversational flexibility \cite{maicher_artificial_2023}. The advent of powerful LLMs like GPT-4 has opened new possibilities, demonstrating strong baseline medical knowledge \cite{nori_capabilities_2023} and impressive capabilities in diagnostic dialogues, even outperforming physicians in some text-based evaluations \cite{tu_towards_2024}. Several studies confirm the feasibility of using LLMs to simulate patients for history taking and other interactions \cite{holderried_generative_2024, potter_enhancing_2024, florez_raspatient_2025, de_mattei_are_2024, cook-etal-2025-virtual, emerson-etal-2025-automated}, often perceived positively by students for enabling safe, repeatable practice \cite{de_mattei_are_2024, borg_creating_2024}.

\subsection{VSP Behavior and Personality}
\noindent However, leveraging LLMs effectively for VSPs requires addressing key challenges identified in the introduction and explored in recent work, particularly controlling LLM behavior to maintain role and personality fidelity and avoid hallucinations \cite{florez_raspatient_2025, borg_creating_2024}. Approaches using knowledge graphs and retrieval-augmented generation show promise for grounding responses in factual data for more reliable simulation \cite{yu2024aipatientsimulatingpatientsehrs, du_llms_2024, li_leveraging_2024}. Simulating diverse personalities and challenging communication styles, beyond just conveying medical information, is another focus, with studies exploring sophisticated prompt engineering to model specific personas \cite{bodonhelyi2025beyond}, although the fundamental limitations of LLMs in replicating deep human states persist \cite{wang_chatgpt_2025}. Our work incorporates personality with Big Five traits, building on these efforts.

\subsection{Vignette generation}
\noindent Automating the creation of complete, diverse and inclusive clinical scenarios is another area where LLMs offer potential \cite{sumpter2024automatedgenerationhighqualitymedical, benoit_chatgpt_2023, bakkum2024using, reichenpfader_simulating_nodate, Faferek2025ChatGPTVirtualPatients}. These studies demonstrate a rapid generation and adaptation of vignettes for different contexts and diversity requirements, using structured prompts and templates. Despite these promising findings, very few studies incorporate these artificial vignettes in their LLM-based conversation generation pipelines. For example, \cite{tu_towards_2024} introduce AMIE, a fine-tuned version of PaLM 2 for diagnostic dialogue, and include a vignette generation module, moderator, dialogue, critic and doctor agent. Additionally,  \cite{lee-etal-2025-adaptive} generate clinically grounded vignettes using Claude 3.5 Sonnet, which are evaluated by domain experts. These vignettes are then used as a foundation for the VSP during conversations. \cite{amithasagaran2025clivrconversationallearningvirtual} adopt a similar approach by integrating artificial vignettes into a conversation pipeline using LLMs. Our generator agent adopts similar principles as these studies, but integrates EBM grounding, and configurable parameters for difficulty and personality.

\subsection{Automated Feedback}
\noindent While the educational aspect of automated feedback is gaining traction, it is often lacking or underdeveloped in VSP literature \cite{kelly2022scoping, bowers_artificial_2024}. Some systems incorporate AI-powered feedback on transcripts \cite{potter_enhancing_2024,borg_creating_2024}, use LLM ensembles on evaluation checklists for assessment \cite{li_leveraging_2024}, or leverage LLMs to generate feedback on limited parts of a consultation, such as history taking in \cite{Bruegge2024LLMDecisionMaking}, or communication using Likert-scale scores in \cite{geathers2025benchmarkinggenerativeaiscoring}. Other studies integrate checklists curated by domain experts in prompts \cite{schiott-etal-2025-using}. Unsupervised coevolution frameworks like EvoPatient also implicitly involve feedback through experience gathering \cite{du_llms_2024}. However, these approaches either lack educational grounding such as evaluation frameworks, rely on checklist conversions that are not available with generated vignettes, or are very narrow in scope.

Relative to these studies, our key contribution lies in presenting a framework that integrates configurable scenario generation, controlled persona-driven interaction, and comprehensive, standards-based automated feedback, addressing limitations throughout the full VSP training lifecycle often tackled piecemeal in previous research.

\section{System architecture}\label{sec:sysarch}
\begin{figure*}[ht]
    \centering
    \includegraphics[width=0.8\linewidth]{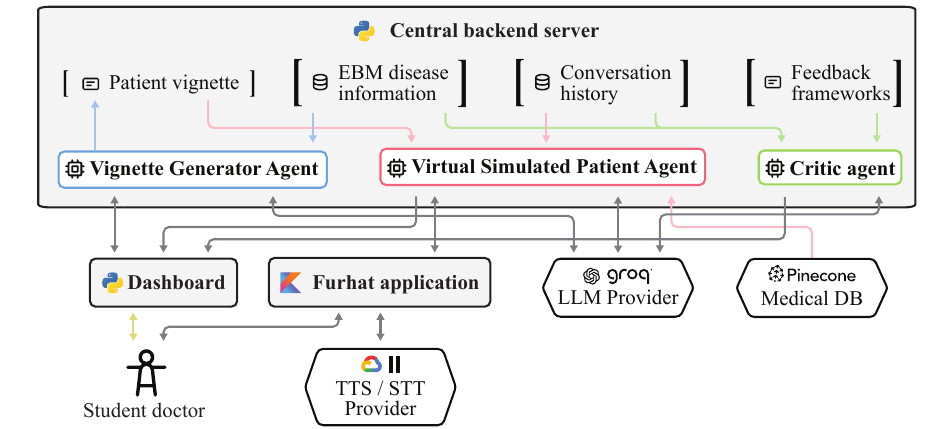}
    \caption{Component diagram with the core contributions from Fig. \ref{fig:maincontributions} to the other components in the framework.}
    \label{fig:simple_system_architecture}
\end{figure*}

\begin{figure}[!t]
    \centering
    \includegraphics[angle=-90,origin=c,width=0.5\linewidth]{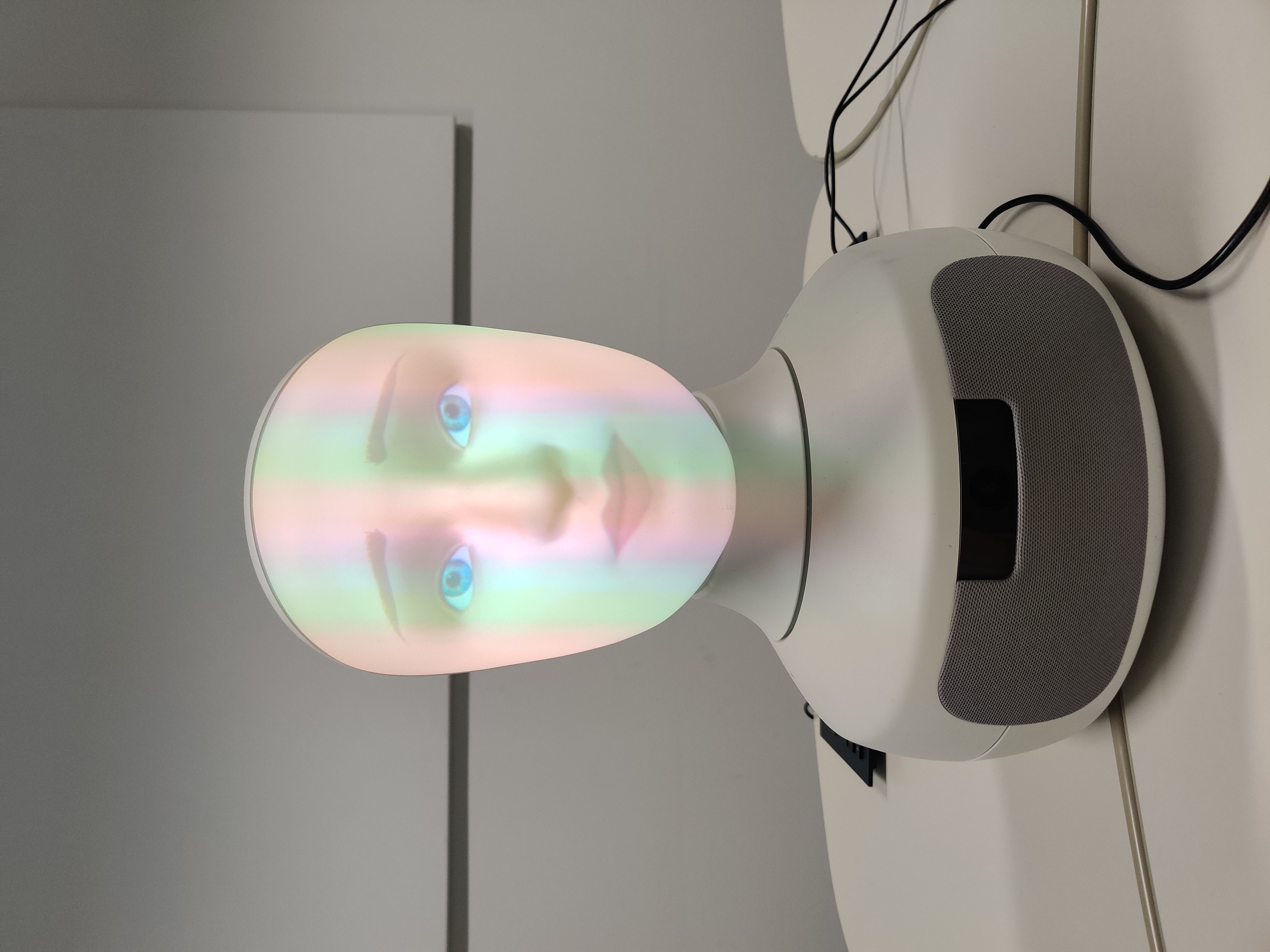}
    \caption{A Furhat robotic head embodying the speech-to-text and text-to-speech of the VSP agent.}
    \label{fig:furhat}
\end{figure}

\noindent The framework uses a distributed, three-tier architecture comprising a web-based client dashboard, a central Python backend server, and an embodied agent application running on the Furhat platform\footnote{[Online]. Available: \url{https://furhatrobotics.com/}} (Fig. \ref{fig:furhat}). An overview can be found in Fig. \ref{fig:simple_system_architecture}, and a more elaborate version in the Appendix.

A training session proceeds as follows. The student configures VSP parameters or selects a predefined case in the dashboard (Section \ref{sec:dashboard}). The generator agent then creates a detailed patient vignette on the backend server (Section \ref{sec:centralbackendserver}). During the consultation, the student interacts through spoken dialogue: the Furhat application handles real-time speech-to-text and text-to-speech I/O (Section \ref{sec:furhatapplication}), while the conversational VSP agent and critic agent run on the backend server. The critic produces both in-session quick tips and post-session standards-based feedback, which is delivered back to the dashboard for review. A full end-to-end example using the framework is provided in the Supplementary Material.

\subsection{Central backend server}\label{sec:centralbackendserver}
\noindent The central backend server, built in Python, acts as the orchestration hub and hosts the core agentic logic. It is responsible for maintaining the overall state of the simulation, including conversation history, patient status, and feedback data. It manages persistent WebSocket connections from both the web client and the Furhat application%
. All three key AI agents are housed on the server: the scenario generator agent, the VSP conversational agent, and the feedback critic agent.

We employ a tiered model architecture to balance medical fidelity against conversational latency. 
For asynchronous tasks requiring high adherence to EBM sources and complex evaluation rubrics (generator and critic agents), we utilize GPT-4.1, prioritizing instruction-following reliability and grounded, structured outputs over runtime. 
Conversely, for the synchronous VSP conversational agent, we prioritize low latency to maintain natural spoken dialogue. We therefore utilize Llama 4 variants for core reasoning and open-weight control, paired with GPT-4o-mini for high-speed, cost-efficient post-processing and formatting.

The full flow of these agents can be found in the Appendix.

\subsubsection{Generator agent (\circnum{1} and \circnum{3})}
The generator agent implements the automated VSP scenario creation. Upon receiving configuration input (desired disease difficulty, Big Five personality scores) from the client, it initiates a generation process. All prompts can be found in the Supplementary Material.
\begin{enumerate}
    \item \textbf{Disease selection and difficulty adjustment:} A disease is picked randomly from a selection of the list of diseases and problems in the educational standards for students finishing the basic medical training at the Faculty of Medicine and Health Sciences in the University of Ghent\footnote{[Online]. Available: \url{https://www.ugent.be/ge/nl/studenten/opleidingsspecifieke-informatie/stages/eindtermen/hoofdstuk3.pdf}}. This represents a realistic Flemish educational context. The difficulty of the entire selection was previously judged by GPT 4.1 on a Likert scale from 1 to 10. Using prompt engineering, complicating or easing factors (e.g., allergies, comorbidities) for the patient vignette are generated based on the evidence-based medicine (EBM) pages pre-coupled with the chosen disease.
    
    \item \textbf{Vignette generation:} The agent prompts GPT-4.1 to generate a detailed patient vignette adhering to a structured template distilled from the University of Tennessee Health Science Center\footnote{[Online]. Available: \url{https://webprod8.uthsc.edu/simulation/resources/case-development.php}}, incorporating modifications suggested in the previous step to match the target difficulty. 
    
    \item \textbf{Consistency check and refinement:} The generated vignette undergoes a final check with GPT-4.1 to identify and correct between elements of the vignette.
    
    \item \textbf{Persona selection:} An appropriate Furhat face and ElevenLabs voice are selected via another GPT-4.1 call, given the vignette content.

    \item \textbf{Personality translation:} Finally, the Big Five personality scores are translated to a textual prompt using the conversion table given in the Appendix for more relevance in the context of patient communication, hereby partially realizing contribution \circnum{3}. These texts are used when personality is included in prompts of the VSP agent.
    
\end{enumerate}
This process ensures scenarios are not only configurable but also grounded in medical data and internally consistent. Therefore, the generation process above fully covers our main contribution \circnum{1}.

\subsubsection{VSP conversational agent (\circnum{2} and \circnum{3})}\label{sec:vspagent}
The VSP conversational agent manages the dialogue flow and implements the controlled response generation. When it receives transcribed user speech (doctor's utterance) from Furhat, it executes a multi-step reasoning pipeline to determine the response:

\begin{figure*}
    \centering
    \includegraphics[width=0.8\linewidth]{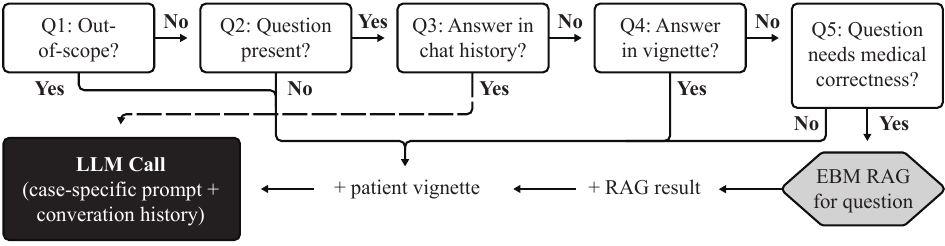}
    \caption{Multi-step pre-processing and RAG steps of the VSP agent before response generation. The corresponding prompt can be found in the Supplementary Material.}
    \label{fig:pre-processing}
\end{figure*}

\begin{enumerate}
    \item \textbf{Utterance preprocessing:} An initial LLM call to LLaMA 4 Scout \cite{llama4scout} analyzes the doctor's last utterance in the context of the conversation history and patient vignette to classify the required response type (e.g., answer directly from vignette, requires external knowledge, question already answered, no question asked). The decision flow can be found in Fig. \ref{fig:pre-processing}, and the full prompt in the Supplementary Material. An exception to this step is the first message, where a custom first message prompt that includes the personality and vignette, is automatically chosen, and the next step is skipped.
    
    \item \textbf{Information 
    retrieval (conditional):} Based on the pre-processing decision, the agent decides on the generation prompt and included information in there (cf. Fig. \ref{fig:pre-processing}; the full prompt can be found in the Supplementary Material). If the answer is likely in the vignette, it uses that information. If external medical knowledge is needed (e.g., typical duration of a symptom not specified in the vignette), it formulates a query and uses a RAG pipeline (LlamaIndex \cite{Liu_LlamaIndex_2022} with a vector database containing EBM articles) to retrieve relevant medical context. If common sense is sufficient, it proceeds without RAG.
    
    \item \textbf{Response generation:} LLaMA 4 Maverick \cite{llama4scout} generates a draft response given the chosen generation prompt supplemented with the retrieved information (this can include vignette and RAG context), the conversation history, and the personality texts.
    
    \item \textbf{Personality and post-processing:} A final LLM call to GPT-4o-mini (prompt in the Supplementary Material) refines the draft response. This critical step ensures the answer adheres strictly to the textual personality constraints, avoids making diagnoses or proposing treatments (unless echoing the doctor), maintains consistency, and fits conversational norms, such as a removal of thoughts, actions, or descriptions beyond the patient's spoken words.
    
\end{enumerate}
This pipeline prioritizes fidelity to the scenario and medical correctness (via RAG and constraints) while incorporating personality and mitigating hallucinations, and therefore covers our main contribution \circnum{2}. The generation step, but mostly the post-processing step, ensure main contribution \circnum{3}.

\subsubsection{Critic agent (\circnum{4})}
The critic agent provides automated standards-based feedback. All prompts can be found in the Supplementary Material.
\begin{itemize}
    \item \textbf{Communication feedback:} 
    Feedback generation occurs in two phases, using the conversation transcript and evaluation frameworks to prompt language models. During the conversation, GPT-4o-mini is prompted based on the best practices of \cite{king2013best} to provide concise, actionable quick tips. After the session, a detailed analysis is performed by prompting GPT-4.1 based on the Master Interview Rating Scale (MIRS) \cite{stillman1976use}. GPT-4.1 returns scores on a 1-5 Likert scale and textual justifications supported by quoted evidence for each MIRS item, generated with a temperature setting of 0.1.
        
    \item \textbf{Clinical feedback:} Post-session, the agent prompts GPT-4.1 to compare the student's implicit diagnostic reasoning and proposed management (extracted from the conversation) against the \textit{gold standard} evidence-based common practice information derived from the EBM sources pre-coupled with the generated disease. It provides structured feedback across predefined clinical categories: diagnosis, treatment planning, follow-up and monitoring, adherence to guidelines, risk assessment, test and investigation ordering, and preventive care.
\end{itemize}
This feedback, corresponding to contribution \circnum{4}, is then sent to the client dashboard via WebSockets.

\subsection{Client dashboard}\label{sec:dashboard}
\begin{figure*}
    \centering
    \includegraphics[width=1\linewidth]{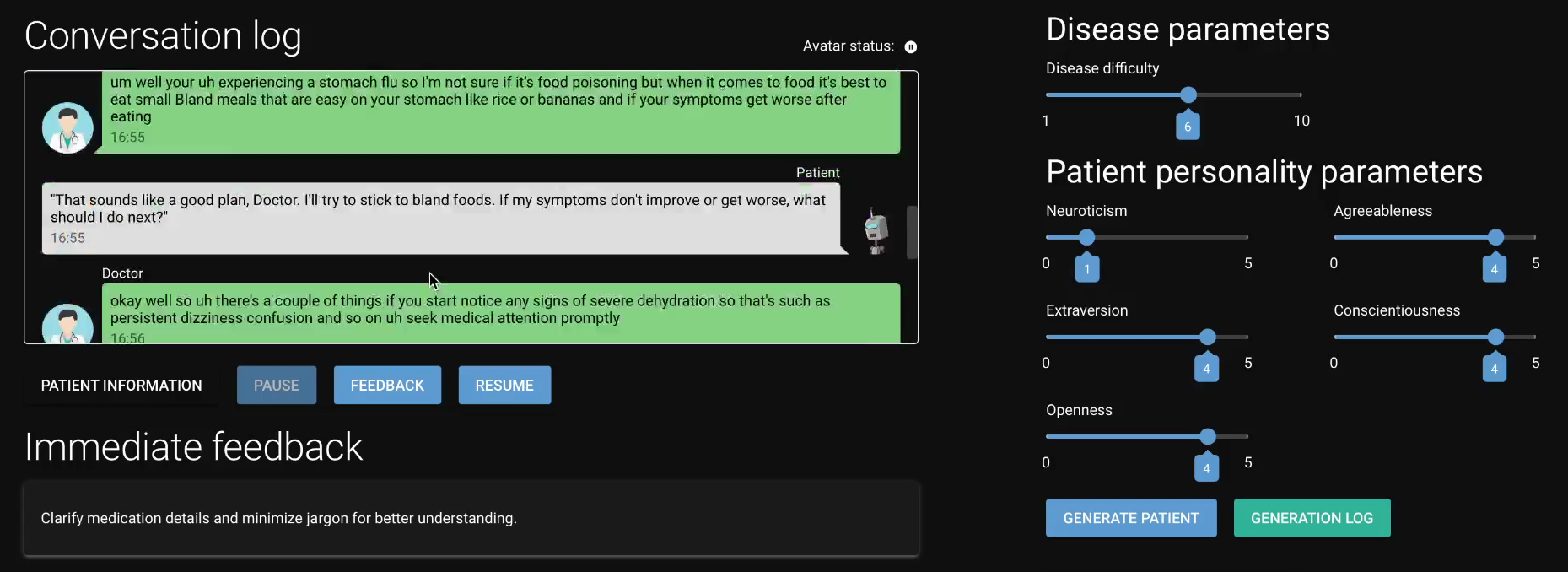}
    \caption{Screenshot of the client dashboard. The left side contains the conversation log, the avatar controls and the immediate feedback. The right side contains the VSP configuration parameters, the patient generation launch button, and the generation log button.}
    \label{fig:dashboard}
\end{figure*}
\noindent The client dashboard (Fig. \ref{fig:dashboard}), developed using the Python NiceGUI framework \cite{nicegui}, serves as the primary user interface. It allows users to configure training scenarios (selecting disease difficulty and personality parameters), monitor the ongoing conversation log, view real-time and final feedback, and control the simulation state (e.g., pause, resume, launch predefined patients). Communication between the client and the server is handled exclusively via WebSockets.

\subsection{Furhat application}\label{sec:furhatapplication}
\noindent The Furhat application, developed in Kotlin, runs on the Furhat robot or its virtual equivalent. Using its WebSocket connection to the central server, its primary roles are real-time speech-to-text (STT) transcription of the user's voice input and text-to-speech (TTS) synthesis of the VSP agent's responses. While the Furhat component handles the physical or virtual embodiment and direct speech I/O, the cognitive processing and decision-making for the VSP's responses reside entirely %
on the backend server.

\section{Evaluation}\label{sec:evaluation}
\subsection{Methodology}
\noindent We evaluated our framework's four main contributions (\textit{cf.} Fig. \ref{fig:maincontributions}) with a user study involving five late undergraduate and nine graduate-level medical students (total $N$=14) at the University of Antwerp (Belgium). Participants provided informed consent and indicated being open to conducting the consultation with a VSP in English. Each evaluation session took place in person, with one user and one instructor present. The evaluation alternated between (i) interacting with the Furhat-based VSP in English and (ii) completing a laptop-based survey in Dutch.

The evaluation comprised four consecutive parts:
\begin{enumerate}
    \item \textbf{Pre-intervention questionnaire}
    \item \textbf{Case 1}: The instructor introduced the dashboard and asked the participant to interact with a VSP (predefined difficulty, personality, and vignette) to find the correct diagnosis. The VSP was loaded into the Furhat and the conversation had no time limit. After ending the interaction and reading the automatically generated feedback, the participant completed a case-specific post-intervention questionnaire. The diagnosis was \textit{acute simple cystitis}.
    \item \textbf{Case 2}: The same procedure was repeated for a second VSP, followed by a case-specific post-intervention questionnaire. The diagnosis was \textit{pancreatitis}.
    \item \textbf{Final post-intervention questionnaire}
\end{enumerate}

The questionnaire was structured in blocks aligned with the procedure: one pre-intervention block, two case-specific post blocks (after each patient), and one final post block. Most questions used 5-point Likert scales and were complemented by open-ended questions. 

The pre-intervention block captured study phase, prior experience with multiple simulation modalities, and baseline perceptions of simulation-based training and AI tools in medical education. 

Each case-specific post questionnaire assessed medical realism, perceived difficulty, perceived inconsistencies, perceived Big Five personality (trait ratings + consistency item), usefulness of quick tips, and perceived clarity/accuracy/usefulness of generated feedback (communication and clinical). 

The final post block assessed overall realism, ease of use, intention to reuse/recommend, practical constraints (e.g., waiting time, embodiment, language), included the System Usability Scale (SUS) \cite{brooke1996sus}, which is a standardized ten-item questionnaire widely used to measure perceived usability of a system, and concluded with open-ended questions about strengths, weaknesses, and integration in the curriculum.

\noindent The full (translated) survey is included in the Supplementary Material.

\subsection{Results and discussion}
\begin{figure}[!t]
    \centering
  \subfloat[VSP 1: target profile (solid) vs student estimate mean (dashed) with ±1 SD band (shaded).\label{1a}]{%
       \includegraphics[width=0.45\linewidth]{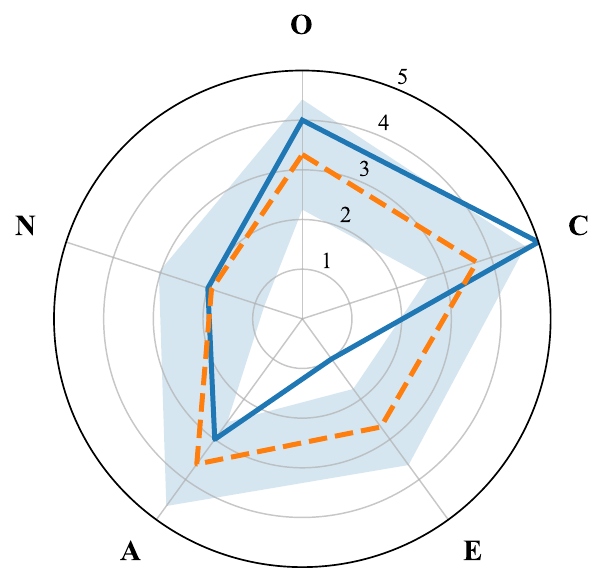}}
    \hfill
  \subfloat[VSP 2: target profile (solid) vs student estimate mean (dashed) with ±1 SD band (shaded).\label{1b}]{%
        \includegraphics[width=0.45\linewidth]{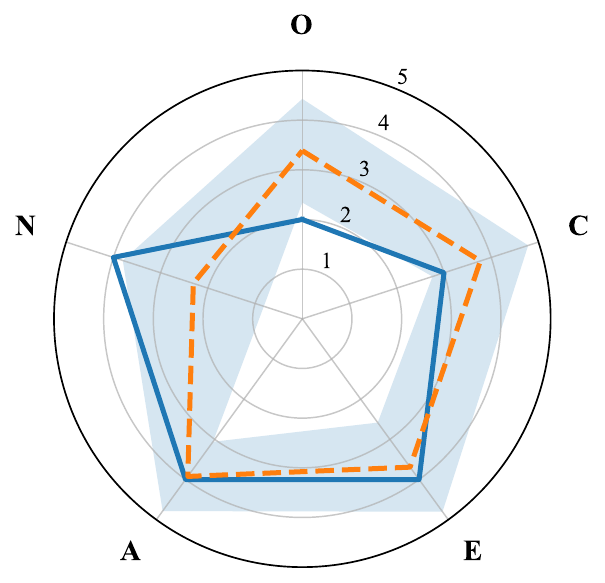}}
    \\
  \subfloat[Comparison of student mean estimates: Person 1 (dashed) vs Person 2 (dotted).\label{1c}]{%
        \includegraphics[width=0.45\linewidth]{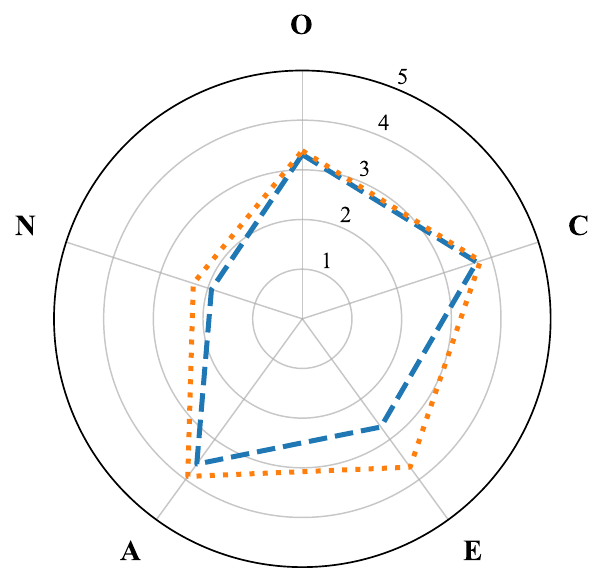}}
    \hfill
  \caption{Radar plots of Big Five personality (1–5 scale) of the two VSP scenarios in the evaluation, $N = 14$. O = Openness, C = Conscientiousness, E = Extraversion, A = Agreeableness, N = Neuroticism.}
  \label{radar_bigfive_evaluation} 
\end{figure}

\begin{table}[!t]
\centering
\small
\caption{Summary of Quantitative Evaluation Outcomes ($N=14$). Values are Mean (SD) and Use a 5-point Likert Scale Unless Otherwise Noted}
\label{tab:results_summary}
\begin{tabularx}{\linewidth}{@{}X c c@{}}
\toprule
\textbf{Measure} & \textbf{Case 1} & \textbf{Case 2} \\
\midrule

\multicolumn{3}{@{}l}{\textbf{Background (reported once)}}\\
Experience with actor-based SPs & \multicolumn{2}{c}{4.21 (0.70)} \\
Experience with virtual SPs & \multicolumn{2}{c}{1.71 (0.91)} \\
Experience with AI-driven virtual SPs & \multicolumn{2}{c}{1.57 (0.94)} \\
\addlinespace

\multicolumn{3}{@{}l}{\textbf{Case-specific outcomes}}\\
Medical realism of VSP answers & 3.93 (0.62) & 3.64 (0.74) \\
Perceived scenario difficulty (0--10) & 4.29 (2.06) & 6.07 (1.89) \\
Personality consistency & 4.21 (0.58) & 3.86 (0.77) \\
Usefulness of quick communication tips & 2.79 (0.70) & 2.79 (0.89) \\
Communication feedback (composite$^{\dagger}$; 5 items) & 4.23 (0.55) & 4.14 (0.74) \\
\quad Includes relevant examples (single item) & 4.43 (0.85) & 4.43 (0.65) \\
Clinical feedback (composite$^{\dagger}$; 4 items) & 4.32 (0.40) & 4.02 (0.69) \\
\addlinespace

\multicolumn{3}{@{}l}{\textbf{Overall evaluation (reported once)}}\\
Overall realism of interactions & \multicolumn{2}{c}{3.36 (0.50)} \\
Interest/engagement & \multicolumn{2}{c}{4.29 (0.61)} \\
Recommend to peers & \multicolumn{2}{c}{4.21 (0.97)} \\
\quad Rated 4 or 5 & \multicolumn{2}{c}{11/14 (79\%)} \\
Reuse for other scenarios & \multicolumn{2}{c}{4.36 (0.74)} \\
\quad Rated 4 or 5 & \multicolumn{2}{c}{12/14 (86\%)} \\
Belief: VSPs prepare for station exam (pre $\rightarrow$ post) & \multicolumn{2}{c}{3.29 (0.47) $\rightarrow$ 4.36 (0.84)} \\
System Usability Scale (SUS; 0--100) & \multicolumn{2}{c}{80.36 (10.37)} \\
\bottomrule
\end{tabularx}
\vspace{0.25em}
\parbox{\linewidth}{\footnotesize $^{\dagger}$Item-level feedback ratings are reported in Table~\ref{tab:feedback_items}.}
\end{table}

\begin{table}[!t]
\centering
\small
\caption{Item-level Ratings of the Automatically Generated Feedback. Values are Mean (SD). Items Use a 5-point Likert Scale}
\label{tab:feedback_items}
\begin{tabularx}{\linewidth}{@{}X c c@{}}
\toprule
\textbf{Item} & \textbf{Case 1} & \textbf{Case 2} \\
\midrule

\multicolumn{3}{@{}l}{\textit{Communication feedback (5 items)}}\\
Clear and easy to understand & 4.29 (0.73) & 3.93 (1.07) \\
Accurately reflected my interaction & 4.07 (0.73) & 4.29 (0.83) \\
Accurately reflected communication guidelines I know & 4.07 (0.83) & 3.93 (1.14) \\
Contained relevant examples from my interaction & 4.43 (0.85) & 4.43 (0.65) \\
Useful to identify areas for improvement & 4.29 (0.73) & 4.14 (1.10) \\
\addlinespace

\multicolumn{3}{@{}l}{\textit{Clinical feedback (4 items)}}\\
Clear and easy to understand & 4.36 (0.63) & 3.79 (1.05) \\
Accurately reflected my interaction & 4.36 (0.63) & 4.21 (0.80) \\
Accurately reflected relevant medical guidelines I know & 4.00 (0.78) & 4.00 (0.96) \\
Useful to identify areas for improvement & 4.57 (0.65) & 4.07 (1.14) \\
\bottomrule
\end{tabularx}
\end{table}

\noindent Table~\ref{tab:results_summary} summarizes the quantitative outcomes.

In the pre-intervention questionnaire, participants reported substantial prior experience with actor-based simulated patients, but limited exposure to virtual simulated patients and especially to AI-driven virtual patients. This suggests that for most participants, this was their first encounter with an AI-based VSP.

In the case-specific questionnaires, students rated the medical realism of both VSPs positively, and no participant reported inconsistencies or incorrect information during the conversations. In an open question about the framework’s strengths, 85\% of students (counting mentions) emphasized the realism of the VSP responses. The perceived difficulty of the interactions was rated close to the generated difficulty of 5/10. This suggests a correct calibration of the generator agent and the VSP conversational agent (\circnum{1} and \circnum{2}).

After reading an informational text about the Big Five traits, participants rated both patients’ personalities. Figure~\ref{radar_bigfive_evaluation} compares the scripted profiles with student estimates. The students’ scores cluster tightly (SD $\approx$ 1; SE $\approx$ 0.30) and yield narrow 95\% CIs ($\pm$ 0.6–0.7). Within those intervals, the two patients show a clear, reproducible profile, most notably a +1.05 point gap in extraversion. Personality consistency was rated high, indicating that the conversations conveyed a stable personality signal. This also corresponds to the answers in an open question on the personality, where 8 out of 14 students indicated the personality to be clear, credible, or realistic (\circnum{3}).

However, although consensus is high, accuracy is limited. For both patients, three of five traits differ from the scripted personality by $\geq$ 0.6 points and were significantly different in one-sample tests ($\lvert t \rvert \geq 2.3$, $p \leq 0.04$). Students therefore agree with one another yet misjudge multiple traits. This pattern has two plausible explanations: (i) the brief, verbal interactions may support fast convergence on similar global impressions while limiting mapping onto accurate Big Five scores (and constraining less observable traits) \cite{blackman1998effect,rouse2003exploring,tskhay2014perceptions}; and (ii) there may be room to further refine how traits are expressed by the VSP conversational agent, for instance using more detailed prompt engineering, few-shot learning or fine-tuning \cite{bodonhelyi2025modeling, kong2024better, brito2025modeling}, or through mechanistic methods such as steering model internals \cite{bhandari2025activation}.

Regarding feedback, students reported high satisfaction with both communication and clinical feedback, with item-level ratings detailed in Table~\ref{tab:feedback_items}. They particularly valued the inclusion of concrete dialogue excerpts in the communication feedback, and they judged the feedback as well-aligned with their interaction and with known communication and clinical guidelines. Clarity was rated favorably overall, but 6/14 students indicated the feedback was too long. The feedback was consistently perceived as useful for identifying areas for improvement. In contrast, the in-conversation quick tips were rated comparatively lower, suggesting they may be distracting when students follow a familiar consultation pattern. Overall, these results suggest that the critic agent reliably delivers clear, guideline-concordant, example-rich commentary perceived as actionable, while leaving room to improve conciseness and reconsider the quick tips’ availability (\circnum{4}).

In the final post-intervention block, overall realism was rated moderately, suggesting a VSP is more a supplement rather than a replacement of real simulated patients. Engagement was rated high and the intention to adopt the framework was likewise strong, with high willingness to recommend it and to reuse it for other scenarios. Moreover, the participants’ belief that VSP interactions would prepare them for their station exam increased from pre- to post-intervention.

Across open-ended responses, students most often highlighted usability and scalability, medical accuracy, scenario realism, and direct feedback. Reported improvement points included processing time (3.4s on average with an SD of 1.84s, excluding outliers above Q3 + 1.5 $\cdot$ IQR, and an additional 3-second listening timeout), and the lack of non-verbal communication on both the receiving and providing end. The SUS score indicated favorable usability \cite{brooke1996sus}.

\section{Conclusion}
\noindent To address existing challenges in virtual simulated patients (VSPs), including realism, consistency, scenario generation for simulation, and comprehensive feedback, we presented an agentic, LLM-based VSP framework to support the full training lifecycle of general practitioner (GP) skills: \textit{configurable scenario creation}, \textit{reliable persona-driven interaction}, and \textit{standards-based assessment and feedback}.

The framework separates concerns across three roles. The \textbf{generator agent} constrains the simulation by producing medically grounded, configurable vignettes from evidence-based medicine (EBM) sources. The \textbf{conversational agent} prioritizes \textit{scenario fidelity} and \textit{medical grounding} through a controlled generation pipeline (structured reasoning, optional retrieval augmentation, and a final constraint/cleanup step), while enabling systematic variation in interaction style via Big Five trait settings. The \textbf{critic agent} links transcript evidence to explicit standards by generating automated communication feedback using MIRS and clinical feedback by comparison against EBM-derived guideline targets. 

While we instantiate these roles with specific models and a spoken interaction setup, the framework is designed to generalize by swapping vignette templates, grounding sources, evaluation frameworks, and underlying language models without changing the overall control structure.

In a student evaluation ($N{=}14$), the VSP answers were rated medically realistic (mean $\approx 3.8/5$), with no reported inconsistencies or incorrect scenario information. Perceived difficulty was close to the targeted mid-level setting (mean $\approx 5/10$ across cases), suggesting effective calibration of scenario generation. Feedback after the conversation was a key strength: both communication and clinical feedback were rated highly (overall $\approx 4.2/5$), and students particularly valued the inclusion of concrete dialogue excerpts to justify feedback. However, always-on in-conversation quick tips may be distracting. Usability and adoption intent were strong (SUS $\approx 80$), with high engagement, high willingness to recommend and reuse the system, and increased belief that VSP interactions can prepare students for station exams.

Beyond these overall ratings, the results suggest three design implications for LLM-based VSPs: (i) \textbf{controlled grounding and structured response control} can yield reliable, vignette-faithful conversations (as reflected by the absence of reported inconsistencies) while still supporting adjustable difficulty; (ii) \textbf{standards-based, example-rich post-session feedback} is perceived as highly actionable, but must balance depth with conciseness; and (iii) \textbf{persona control} can produce a stable and clearly perceived personality signal, but the current setup motivates calibrated persona controllers with behavior-based evaluation, ensuring that intended trait settings translate into consistently recognizable interaction patterns. General key areas for improvement are reducing processing time and extending communicative realism through non-verbal behavior.

\section{Availability and Licensing}
\noindent The source code is available at: \url{https://github.com/Victordmz/agentic-framework-gp-skills}.

\section{Limitations and Future Work}
\noindent While our framework effectively maintains personality fidelity, medical accuracy in patient simulation, grounded vignette generation, and accurate feedback, several areas for improvement remain.

First, the framework currently lacks integration of non-verbal communication. It neither captures non-verbal cues from the doctor nor allows the VSP to exhibit non-verbal cues toward the doctor, limiting communicative realism beyond verbal interactions, and hindring empathy, impacting communication. Future collaboration with Human-Computer Interaction (HCI) experts could address this gap.

Additionally, we must address specific challenges observed during user studies, such as overly extensive feedback, distracting quick tips, and and the absence of physical examination capabilities. We must also further investigate if there is an inaccurate reflection of certain personality trait combinations in conversations.

Further enhancements are needed in the three agents. The generator agent can be improved with an ability to create longitudinal cases, adhere more closely to established best practices for OSCE case generation, expanding or adapting patient vignette templates, and more systematically defining the difficulty levels of cases. The VSP conversational agent should leverage recent research on personality integration (e.g., \cite{bodonhelyi2025modeling, kong2024better, brito2025modeling, bhandari2025activation}). Reducing generation latency is also desirable. The critic agent could further benefit by incorporating frameworks specifically tailored to various educational contexts.

\section*{Ethical Considerations}
\noindent Training with VSPs should supplement, not replace, human interaction and feedback from real educators and SPs where feasible. The system should not be used for real diagnosis or treatment advice. Ensuring fairness and avoiding harmful stereotypes in generated scenarios and personality portrayals is crucial and requires ongoing monitoring. Transparency about the system being an AI is important for users.

\appendices
\section{Full system architecture and flow}\label{app:fullarch}
\begin{figure*}[!t]
    \centering
    \includegraphics[width=0.9\linewidth]{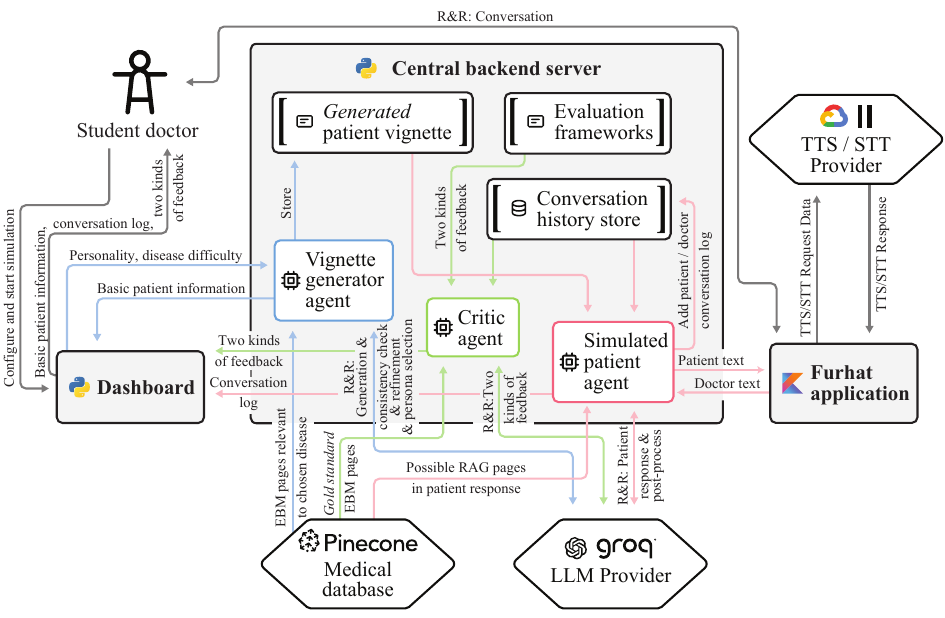}
    \caption{System architecture diagram showing the three core agents (Generator, VSP, Critic) hosted on the server, the client dashboard, the embodied Furhat application, the student doctor user, and key data components (medical DB, vignette, conversation history, evaluation frameworks). Arrows indicate the primary flow of information. \textit{R\&R} indicates a request and response that both convey information.}
    \label{fig:system_architecture}
\end{figure*}

\begin{figure*}
    \centering
    \includegraphics[width=0.9\linewidth]{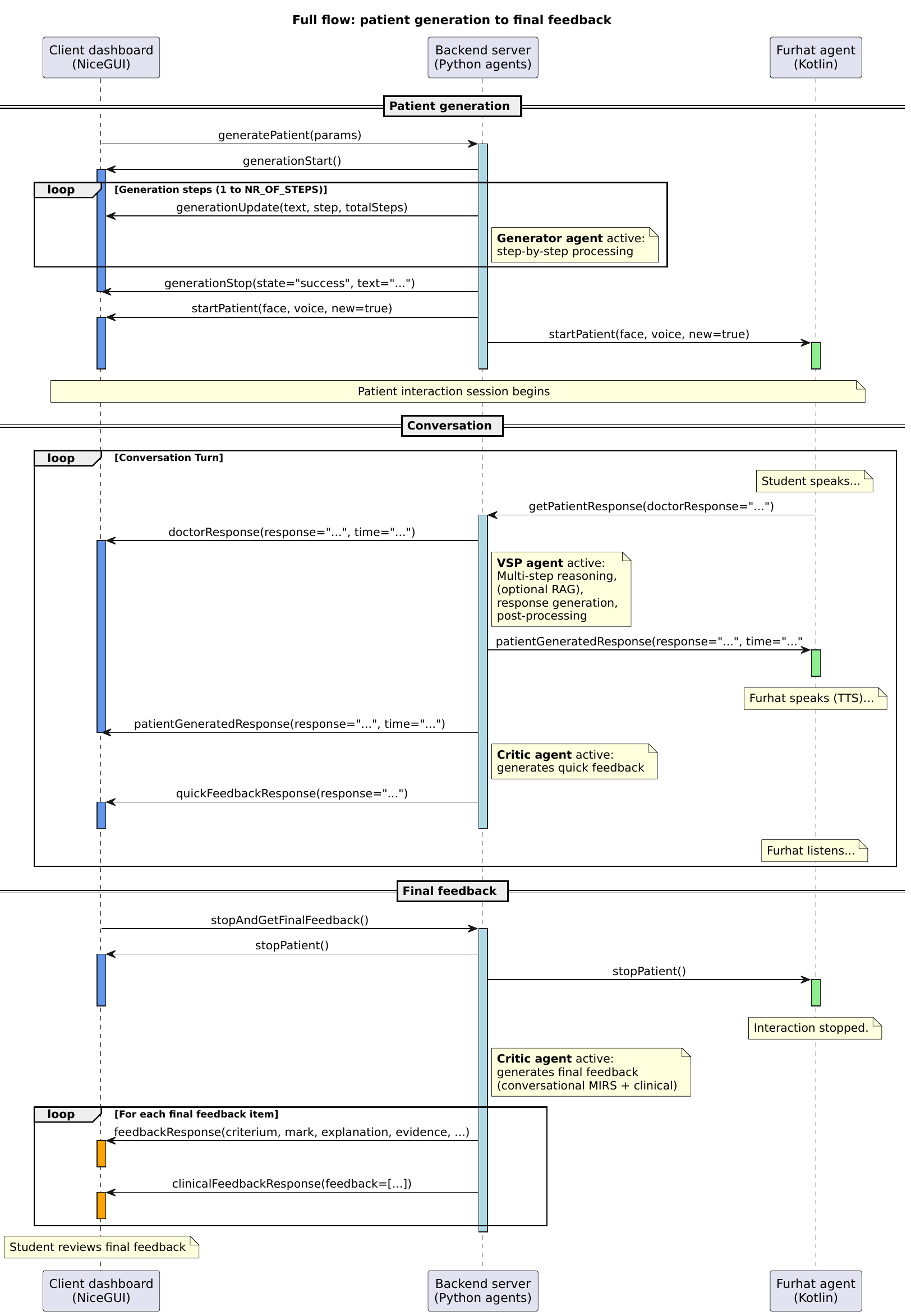}
    \caption{Sequence diagram of the full flow of the framework from generation to final feedback. }
    \label{fig:full-seq-diagram}
\end{figure*}

The full system architecture can be found in Fig. \ref{fig:system_architecture}. The flow of the framework, i.e., the way the system components interact, can be found in Fig. \ref{fig:full-seq-diagram}.

\section{Personality score to text conversion table}\label{app:personalityconversion}

\begin{itemize}

    \item \textbf{Neuroticism}
        \begin{description}
            \item[0:] Consistently discusses even serious symptoms with a calm, detached tone, rarely expressing any worry or seeking emotional reassurance
            \item[1:] Usually maintains a calm demeanor when describing symptoms, seldom expressing anxiety and typically focusing on factual aspects
            \item[2:] Occasionally mentions mild worry or concern about symptoms but quickly pivots back to practical questions, generally remaining composed
            \item[3:] Expresses a moderate level of worry appropriate to the health concern, asking questions seeking both information and reasonable reassurance
            \item[4:] Often expresses significant anxiety about symptoms or potential outcomes, frequently asking 'what if' questions and seeking repeated reassurance from the doctor
            \item[5:] Consistently voices strong fears and worries about their health, often catastrophizing minor symptoms and persistently seeking reassurance
        \end{description}

    \item \textbf{Extraversion}
        \begin{description}
            \item[0:] Speaks only in minimal, quiet, one-or-two-word responses when directly questioned, never initiating conversation or small talk
            \item[1:] Rarely offers information beyond direct, brief answers, speaking softly and infrequently making unsolicited comments
            \item[2:] Answers questions politely but succinctly, seldom elaborating unprompted and maintaining a noticeably reserved, quiet demeanor
            \item[3:] Engages in polite back-and-forth, answers questions reasonably fully, and might offer a brief, relevant personal comment occasionally
            \item[4:] Often elaborates extensively on answers, readily initiates small talk or shares personal anecdotes, speaking with noticeable energy
            \item[5:] Consistently dominates the conversation with lengthy, energetic explanations and frequent personal stories, often filling any potential silences
        \end{description}

    \item \textbf{Openness}
        \begin{description}
            \item[0:] Consistently dismisses any non-standard treatment options mentioned and asks only about the most practical, established procedures
            \item[1:] Frequently steers the conversation back to concrete symptoms and immediate practical steps, rarely asking 'why' things work
            \item[2:] Occasionally asks a clarifying question about the basic mechanisms but primarily focuses on routine and known procedures
            \item[3:] Asks standard questions about the diagnosis and treatment plan, accepting information straightforwardly without much speculation or resistance
            \item[4:] Often asks speculative 'what if' questions about their condition or expresses curiosity about the underlying biological processes involved
            \item[5:] Consistently brings up alternative therapies or research they've read, eagerly exploring various theoretical possibilities for their condition
        \end{description}

    \item \textbf{Agreeableness}
        \begin{description}
            \item[0:] Consistently voices suspicion about the diagnosis or doctor's motives, frequently using challenging or critical language towards recommendations
            \item[1:] Often questions the doctor's suggestions or expertise, expressing skepticism about the necessity or effectiveness of his utterances
            \item[2:] Occasionally voices mild disagreement or doubt about a recommendation, asking probing questions before reluctantly agreeing
            \item[3:] Generally cooperates with the doctor's requests and asks questions politely, expressing concerns in a non-confrontational manner
            \item[4:] Often expresses explicit trust and gratitude towards the doctor, readily agreeing with suggestions with minimal questioning
            \item[5:] Consistently defers to the doctor's judgment with strong verbal agreement, frequently offering praise and avoiding any hint of conflict
        \end{description}

    \item \textbf{Conscientiousness}
        \begin{description}
            \item[0:] Consistently gives vague, disorganized accounts of symptoms and frequently mentions forgetting instructions or medication doses
            \item[1:] Often struggles to recall specific details like symptom timelines or medication names, needing frequent prompting from the doctor
            \item[2:] Sometimes provides incomplete information or needs reminders about previous advice, occasionally mentioning difficulties sticking to the plan
            \item[3:] Provides a reasonably clear account of their main issues and generally affirms understanding of instructions, asking a few basic clarifying questions
            \item[4:] Often comes prepared with details or a mental list of questions and frequently asks for detailed clarification on treatment instructions to ensure accuracy
            \item[5:] Consistently presents details about symptoms and medication adherence, meticulously double-checking every aspect of the treatment plan
        \end{description}

\end{itemize}
\clearpage
\bibliographystyle{IEEEtran}
\bibliography{custom} %
\end{document}